\documentclass[11pt]{article}

\usepackage[final]{acl}

\usepackage{times}
\usepackage{latexsym}
\usepackage{times}
\usepackage{amsmath}
\usepackage{latexsym}
\usepackage{booktabs}
\usepackage{tabularx}
\usepackage{subcaption}
\usepackage{comment}
\usepackage{amssymb}
\usepackage{natbib}
\usepackage[T1]{fontenc}
\usepackage[utf8]{inputenc}

\usepackage{microtype}

\usepackage{inconsolata}

\usepackage{graphicx}
\usepackage{inconsolata}

\usepackage{placeins}

\usepackage{algorithm}
\usepackage{algpseudocode} % Use algpseudocode instead of algorithmic
\usepackage{amsmath} % For math equations
\usepackage{enumitem}
\usepackage{xcolor}
\usepackage{tcolorbox}

\usepackage{tabularx}
\usepackage{colortbl}

\usepackage{booktabs}
\newcommand{\method}{Jev-Mem}

\title{Jev-Mem: System-One-Controlled Agentic Memory for Efficient AI Agents}

\author{
Dongming Jiang, Yi Li, Bingzhe Li\thanks{Corresponding author} \\
Department of Computer Science, The University of Texas at Dallas \\
\texttt{\{dongming.jiang, yi.li3, bingzhe.li\}@utdallas.edu} 
}

\begin{document}
\maketitle
\begin{abstract}
Agentic memory is becoming essential for long-horizon AI agents, yet many existing systems rely on autoregressive LLMs to control how memories are organized, retrieved, and used, placing expensive generation on the critical path of memory operations. We introduce \textbf{\method}, a new agentic memory architecture inspired by System-One/System-Two cognition. System One captures fast, lightweight decision-making, whereas System Two performs slower, deliberative reasoning. Jev-Mem brings this division of labor to agentic memory through a dedicated System-One control plane, a structured multi-relational memory plane, and a System-Two reasoning plane. The System-One controller governs memory typing and relational organization during construction, and dynamically performs query routing, retrieval-budget allocation, graph traversal, candidate scoring, and adaptive stopping during retrieval. System Two is invoked only for complex reasoning and answer synthesis. This design improves both memory effectiveness and system efficiency: on LoCoMo Jev-Mem achieves an overall LLM-as-a-Judge score of 0.777, an 11.0\% relative improvement over the strongest baseline, while reducing memory construction time to 158\,s, a 6.6$\times$ speedup over the fastest competing memory system, and lowering average query latency to 0.93\,s, a 36.7\% reduction. The code of Jev-Mem is publicly available.\footnote{\url{https://github.com/libingzheren/Jev-Mem}}

\end{abstract}

\section{Introduction}
Large language model (LLM) agents are increasingly expected to operate as persistent systems over long interaction horizons, supporting applications such as coding assistants, personal agents, research agents, and autonomous workflows~\citep{Brown2020, achiam2023gpt, Wei2022}. Such agents typically interleave reasoning with tool invocation and environment actions~\citep{yao2023react, schick2023toolformer}, and operate over multi-step environments such as interactive web platforms and software repositories~\citep{zhou2024webarena, yang2024sweagent}. During these interactions, agents continuously accumulate user preferences, task history, and environment knowledge, quickly exceeding what can be maintained within a fixed context window~\citep{Beltagy2020, liu2024lost, Press2021}. Enlarging the nominal context length does not by itself resolve this problem, since models do not reliably exploit all positions of a long input~\citep{hsieh2024ruler}, and long-term interactive settings introduce additional indexing, retrieval, and reading challenges~\citep{lee2024human, wu2024longmemeval, hu2025memoryagentbench}. To remain effective over time, agents therefore need mechanisms that can retain useful experience beyond the prompt and recover it when needed. This requirement has made \emph{agentic memory} which has the ability to preserve, organize, update, and retrieve past experience~\citep{xu2025mem, nan2025nemori, chhikara2025mem0, jiang2026magma, liu2026simplemem, jiang2026anatomy}.

Recent work has transformed agentic memory from passive storage into an active and structured component of LLM agents. Early systems mainly store past interactions and retrieve them through semantic similarity, while newer approaches selectively retain salient information, consolidate repeated observations, and organize memory across hierarchical stores. Procedural memory further captures reusable skills, distilled experience, and recurring workflows~\citep{wang2023voyager, zhao2024expel, wang2024awm}. More recent relational approaches represent memories through knowledge graphs or multiple graph views, enabling retrieval over semantic, temporal, causal, and entity relationships rather than isolated memory fragments.
As memory becomes richer, however, controlling it becomes increasingly expensive. Persistent agents must repeatedly decide what to store, update, connect, retrieve, and when to stop searching. Existing systems typically rely on either fixed heuristics or general-purpose autoregressive LLMs: the former are efficient but inflexible, while the latter provide semantic flexibility at the cost of repeated token generation. As these decisions appear throughout memory construction and retrieval, memory control itself can become a major source of latency and inference overhead.

This observation leads us to reconsider agentic memory through the lens of \emph{System One and System Two}, a distinction drawn from dual-process accounts of human reasoning that separate fast, automatic processing from slower and more deliberative thought~\citep{evans2008dual}. Many memory operations including memory typing, relation judgment, query routing, candidate scoring, and stopping are structured, high-frequency decisions that naturally fit lightweight System-One computation, while evidence synthesis and final answer generation remain better suited to System Two.
Jev~\cite{jev} makes such a separation practical by producing typed probabilistic decisions without autoregressive generation. Yet several questions remain: \emph{How should System One be integrated across the memory lifecycle? How can a weaker reasoner control memory without hurting downstream accuracy? What memory structure and retrieval process best support lightweight control?}

Inspired by this observation, we propose \textbf{Jev-Mem}, a new System-One/System-Two architecture that treats memory control itself as a first-class systems layer. Unlike existing agentic memory systems that rely on heuristics or repeatedly invoke autoregressive LLMs throughout memory construction and retrieval, Jev-Mem introduces a dedicated System-One control plane that handles high-frequency structured decisions, a shared structured memory data plane, and a System-Two reasoning plane reserved for complex synthesis and answer generation. This separation redesigns the full memory lifecycle: on the write path, System One performs memory typing, redundancy filtering, and semantic, temporal, causal, and entity relation construction; on the read path, it performs query routing, retrieval-budget allocation, graph traversal, candidate scoring, evidence assessment, and adaptive stopping. By moving these frequent decisions out of the generative reasoning loop, Jev-Mem reduces unnecessary autoregressive inference while still preserving strong reasoning capability through System Two. Jev provides one concrete realization of the System-One controller through typed probabilistic decisions, but the key novelty of Jev-Mem is the architectural separation of lightweight memory control from deliberative reasoning across both memory construction and retrieval.

We make four main contributions.
\begin{itemize}
    \item We identify an opportunity to leverage System-One-style lightweight decision-making to enable more efficient agentic memory.

    \item We introduce \textbf{Jev-Mem}, a System-One/System-Two architecture that separates lightweight structured memory control from expensive generative reasoning.

    \item We develop a unified System-One control plane for both memory construction and retrieval, including multi-relational organization, query routing, adaptive graph traversal, candidate scoring, and stopping.

    \item Jev-Mem achieves the best overall accuracy while also delivering the highest efficiency among state-of-the-art baselines.
\end{itemize}

\begin{figure}[t]
    \centering
    \includegraphics[width=0.45\textwidth]{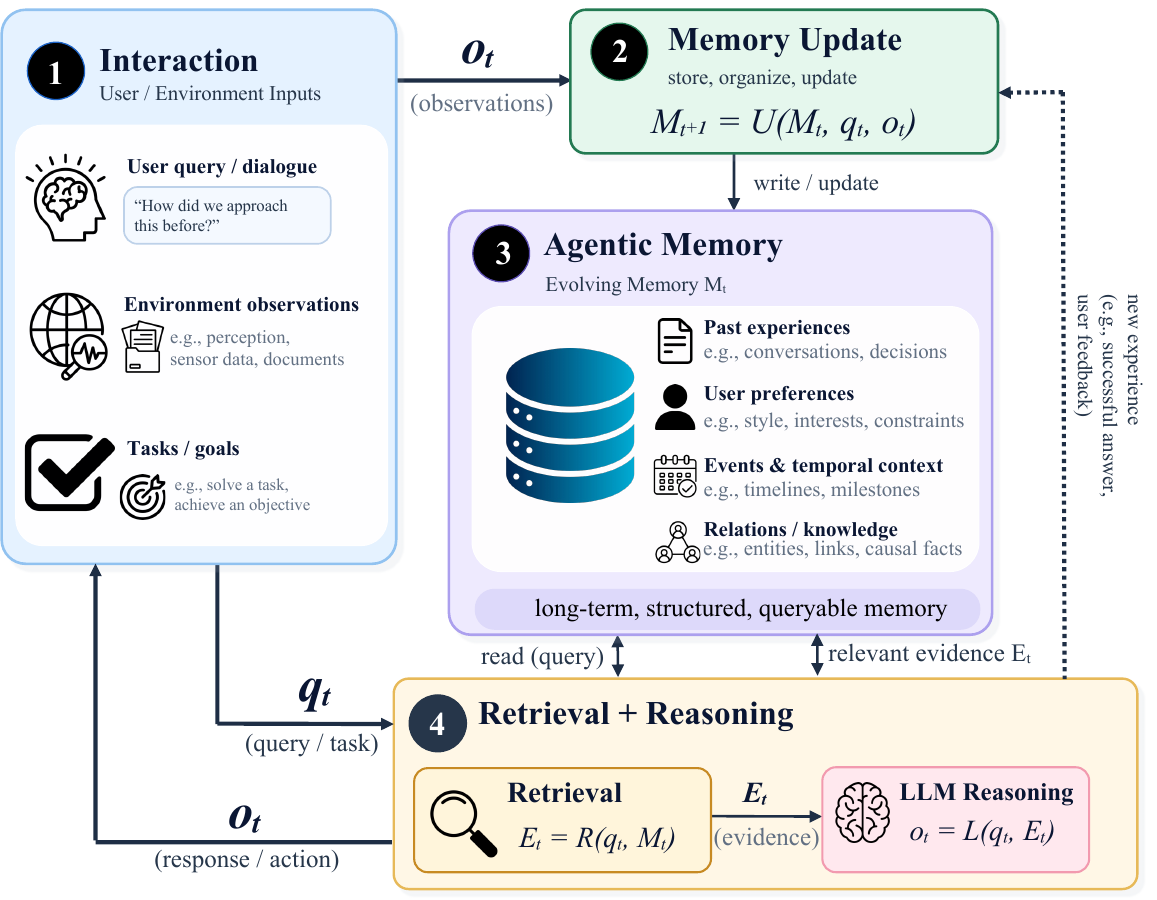}
    \caption{
    Workflow of agentic memory.
    }
    \label{fig:agent_mem}
\end{figure}
\section{Background and Motivation}
\label{sec:background}
\subsection{Agentic Memory}
Figure~\ref{fig:agent_mem} demonstrates the workflow of agnetic memory. Let an agent maintain an evolving memory $M_t$. At interaction step $t$, a query $q_t$ retrieves relevant evidence

\begin{equation}
E_t = R(q_t, M_t)
\end{equation}

which is then provided to a language model for reasoning and response generation:

\begin{equation}
o_t = L(q_t, E_t)
\end{equation}

The resulting interaction may subsequently update the memory:

\begin{equation}
M_{t+1} = U(M_t, q_t, o_t)
\end{equation}

Agentic memory has evolved from simple retrieval over stored interaction histories toward increasingly active memory management. Modern systems may selectively organize observations, infer relationships among memories, consolidate information, route queries across different memory structures, and adapt retrieval based on the current query. As a result, memory is no longer only a passive store accessed by a retrieval function. It increasingly behaves as a dynamic subsystem that continuously makes decisions about how information should be organized and accessed.

This shift introduces an important systems question that has received comparatively less attention: \emph{what computational mechanism should execute these memory-management decisions?} Both memory update $U$ and retrieval $R$ contain frequent semantic decisions. During memory construction, the system must determine how new information should be characterized and connected to existing knowledge. During retrieval, it must determine where to search, which candidates are useful, how much additional search is warranted, and when sufficient evidence has been collected.

External retrieval has long served as a mechanism for augmenting parametric language-model knowledge with non-parametric stores, through dense retrievers, retrieval-augmented pre-training, and retrieval-augmented generation~\citep{karpukhin2020dpr, guu2020realm, lewis2020retrieval, borgeaud2022retro, izacard2023atlas}. Agentic memory inherits much of this machinery, but differs in that its store is written by the agent's own interaction history and must be maintained and reorganized over time rather than fixed in advance.

A closely related line of work studies adaptive control in retrieval-augmented generation. Rather than applying the same retrieval procedure to every query, active and adaptive methods dynamically determine whether retrieval is needed, when additional evidence should be acquired, or which retrieval strategy a given query warrants~\citep{trivedi2023ircot, jiang2023flare, asai2024selfrag, jeong2024adaptiverag}. These results motivate viewing retrieval as a controlled decision process rather than a fixed top-$k$ operation. Jev-Mem builds on this view, and extends the same principle from evidence acquisition to the broader memory lifecycle, including memory construction, query routing, budget allocation, graph traversal, and stopping.

Jev-Mem focuses on this memory-control layer. Rather than treating these decisions as incidental components embedded inside prompts or fixed heuristics, we make them an explicit part of the memory architecture.

\subsection{The Opportunity for System-One Memory Control}
A key observation is that many memory-control operations are \emph{semantic but not generative}. During memory construction, the system may need to classify a memory or infer its relation to existing information; during retrieval, it may need to route a query, score candidates, or decide when to stop. These decisions typically produce bounded outputs such as labels, probabilities, or scores rather than free-form text.

Using an autoregressive LLM for such high-frequency decisions is therefore unnecessarily expensive: even simple judgments require token-by-token generation, formatting, and parsing. Because these operations repeatedly appear on the memory critical path, their overhead can accumulate quickly.

More broadly, cost-aware LLM inference systems have shown that different requests need not receive identical amounts of model computation. Cascading and routing approaches dynamically allocate requests across models of differing cost to improve the cost--quality tradeoff~\citep{chen2023frugalgpt, ong2024routellm}, and speculative decoding uses a small model to draft tokens that a larger model only verifies~\citep{leviathan2023speculative}. Jev-Mem applies a related systems principle at a finer granularity: rather than routing only complete user requests between models, it separates frequent, bounded memory-control decisions from open-ended generative and deliberative reasoning.

This creates a natural opportunity for a System-One/System-Two design: use lightweight structured prediction for frequent memory-control decisions, while reserving System Two for complex reasoning and answer synthesis.

\begin{figure*}[t]
    \centering
    \includegraphics[width=0.8\textwidth]{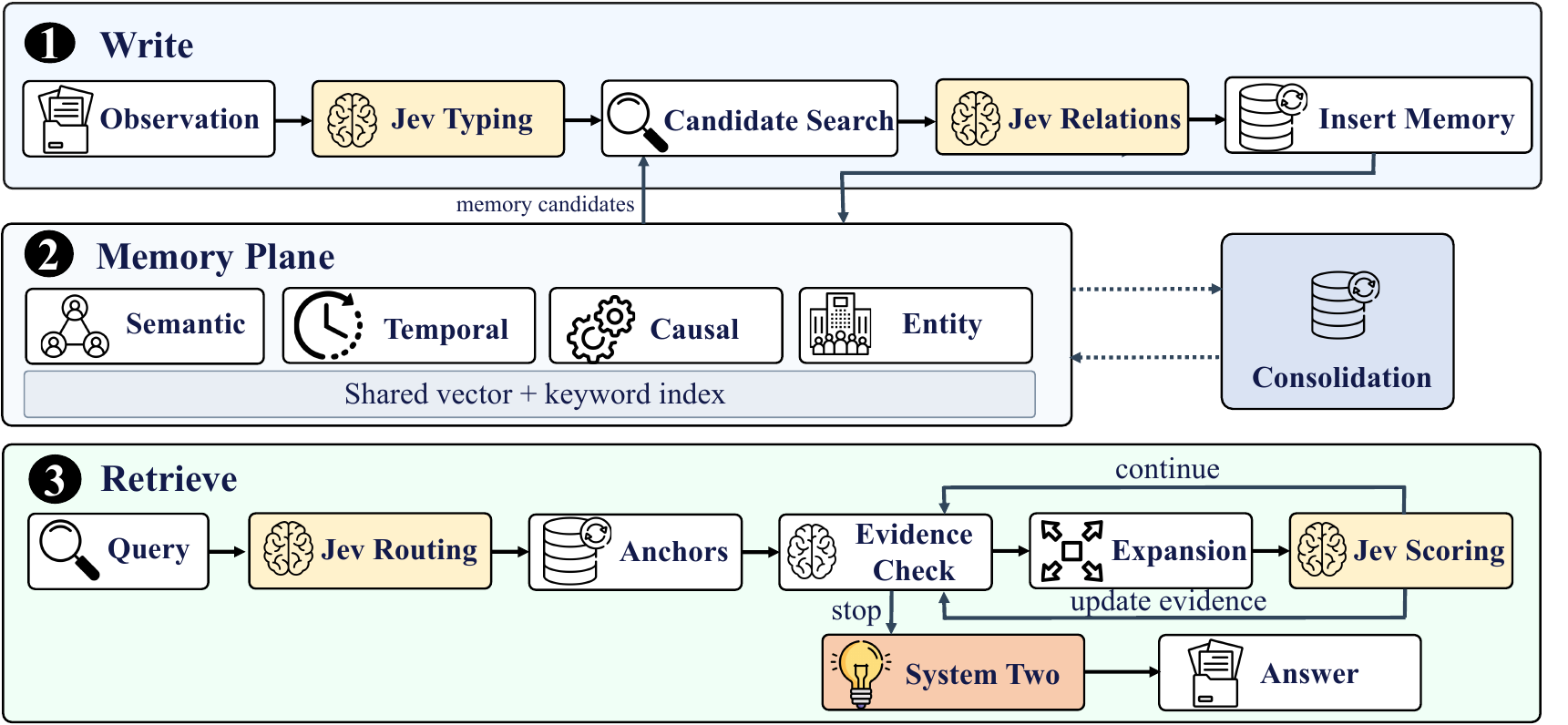}
    \caption{
    Overview of Jev-Mem. A lightweight System-One control plane manages
    memory construction and adaptive retrieval, while System Two is reserved
    for complex reasoning and answer synthesis.
    }
    \label{fig:jev_mem_overview}
\end{figure*}
\section{Jev-Mem Algorithm Design}
\label{sec:algorithm}
Jev-Mem redesigns agentic memory around a clear separation between fast memory control and deliberative reasoning. As illustrated in Figure~\ref{fig:jev_mem_overview}, the architecture consists of three components: a System-One controller, a shared memory data plane, and a System-Two reasoning model. Rather than relying on an autoregressive LLM to manage every memory operation, the System-One controller makes lightweight, structured decisions throughout both writing and retrieval. The memory plane maintains canonical observations together with semantic, temporal, causal, and entity relations, while System Two is reserved for answer synthesis and other operations that require open-ended generation or deeper reasoning.

This architecture unifies memory construction and retrieval under the same control mechanism. On the write path, Jev-Mem preserves observations, assigns memory types, selects candidate memories, and determines how new observations should connect to the existing multi-relational memory. On the read path, it routes each query to the most relevant relational views, allocates retrieval effort, checks evidence sufficiency, expands the graph when necessary, and scores newly discovered candidates in an iterative feedback loop. By treating both workflows as coordinated System-One control processes, Jev-Mem replaces a collection of isolated heuristics and repeated LLM calls with a single, adaptive memory-management architecture.

Let an observation be
\begin{equation}
o_t=(x_t,\tau_t,\mu_t)
\end{equation}
where $x_t$ denotes its content, $\tau_t$ an optional timestamp, and $\mu_t$ its provenance. Jev-Mem maintains
\begin{equation}
\mathcal{M}_t =
\left(
V_t,
\{E_t^g\}_{g\in\mathcal{G}},
I_t^{\mathrm{vec}},
I_t^{\mathrm{lex}}
\right)
\end{equation}
where
\begin{equation}
\mathcal{G}=
\{\mathrm{semantic},
\mathrm{temporal},
\mathrm{causal},
\mathrm{entity}\}
\end{equation}
All relational views share the same canonical memory nodes $V_t$; multiple typed edges may connect the same pair of memories, while the vector and lexical indexes provide complementary entry points into the same memory space.

\subsection{Typed System-One Memory Control}
\label{sec:typed-control}

The central abstraction of Jev-Mem is a typed System-One controller $\mathcal{J}(S,\mathcal{Q})$,
where $S$ is structured state and $\mathcal{Q}$ is a batch of explicit decision questions. The controller produces either probabilities for independent propositions or a distribution over mutually exclusive alternatives. For example, it can estimate whether two memories are semantically related, whether a candidate is relevant to the current query, or whether the retrieved evidence is sufficient. When exactly one action is required, it selects among predefined alternatives such as temporal relations.

This interface is intentionally different from free-form LLM prompting. Each decision exposes a small, known output space, allowing Jev-Mem to represent memory control directly as probabilities rather than generating intermediate natural-language reasoning and subsequently parsing it. Moreover, decisions sharing the same state can be evaluated together in a single batched invocation.

The same interface governs both the write and read paths. On the write path, it controls memory typing and relation construction. On the read path, it performs query routing, candidate evaluation, and stopping. This shared control plane is a key architectural property of Jev-Mem: memory is not constructed by one collection of heuristics and retrieved by another, but instead managed throughout its lifecycle through the same System-One decision abstraction.

\subsection{System-One-Guided Memory Construction}
\label{sec:memory-construction}

For each valid observation, Jev-Mem creates one canonical memory node and selectively constructs relations around it. The current design preserves observations rather than making an irreversible learned store-or-discard decision at ingestion time. This prevents information that appears unimportant initially from being permanently lost before a future query reveals its relevance. Selectivity is instead introduced when memory structure is constructed and later when memory is retrieved.

The controller first predicts four overlapping memory characteristics:
\begin{equation}
\mathbf{t}(v)=
(t_{\mathrm{episodic}},
 t_{\mathrm{semantic}},
 t_{\mathrm{procedural}},
 t_{\mathrm{preference}})
\end{equation}
These scores annotate the node rather than assigning it to a single mutually exclusive category. The canonical node retains the original observation, provenance, timestamp, embedding, entities, and type scores.

A central challenge is determining how a new observation relates to an increasingly large memory. Comparing it against every existing node would make controller cost grow directly with memory size. Jev-Mem therefore separates \emph{candidate discovery} from \emph{relation judgment}. Deterministic retrieval first combines vector similarity, lexical overlap, shared entities, and temporal proximity to identify at most $K_w$ candidates:
\begin{equation}
C(v)=
\operatorname{TopK}_{u\in V_t}
s_{\mathrm{cand}}(v,u)
\end{equation}
The System-One controller then evaluates only these candidate pairs.

For every pair $(v,u)$, Jev estimates semantic relatedness, directional causal influence, same-episode membership, and, when necessary, entity equivalence. Whenever reliable structured information is already available, Jev-Mem avoids unnecessary learned inference. Timestamp ordering directly creates temporal relations, while exact shared identifiers directly create entity relations. When temporal order is implicit, the controller instead chooses among before, after, during, contains, overlaps, same\_time, and unknown.

An inferred edge of type $g$ is inserted only when
\begin{equation}
P(g\mid v,u)\geq\theta_{\mathrm{rel}}
\end{equation}
Because relations are independent views over a common memory space, the same pair of nodes may simultaneously exhibit semantic, temporal, causal, and entity relationships. This differs from assigning each memory to a single graph or duplicating the same observation across several memory stores. The write path is
\begin{equation}
o_t
\rightarrow \mathrm{type}
\rightarrow \mathrm{candidates}
\rightarrow \mathrm{relations}
\rightarrow \mathcal{M}_t
\end{equation}

This design improves memory quality in two ways. First, preserving canonical observations avoids premature information loss. Second, selective relation construction suppresses unnecessary graph connectivity while retaining relations that can later support semantic, temporal, causal, or entity-based reasoning.

Periodic maintenance further examines a bounded neighborhood for redundancy, contradiction, obsolescence, and useful additional links. These decisions enrich the memory structure without removing the original evidence. If a higher-level textual abstraction is required, the control plane may explicitly escalate an approved merge or promotion to System Two; generation is therefore an optional consequence of a structured control decision rather than the default mechanism for memory maintenance.

\subsection{Adaptive System-One Retrieval}
\label{sec:adaptive-retrieval}

Jev-Mem treats retrieval as a closed-loop control process rather than a single top-$k$ search. Given a query $q$, the controller first predicts the relevance of each relational view,
\begin{equation}
\mathbf{p}(q)
=
\{p_g(q)\}_{g\in\mathcal{G}}
\end{equation}
together with a multi-hop requirement $h(q)$ and a recency importance score $r(q)$. Because these probabilities are evaluated independently, a query may activate several graph views simultaneously instead of being assigned to one discrete retrieval intent.

A relation type is active when
\begin{equation}
p_g(q)\geq\theta_{\mathrm{act}}
\end{equation}
Given a total graph-expansion budget $B$, Jev-Mem distributes search effort according to the predicted graph needs:
\begin{equation}
w_g(q)=
\frac{p_g(q)^\gamma}
{\sum_{j\in\mathcal{A}(q)}p_j(q)^\gamma}
\end{equation}
where $\mathcal{A}(q)$ denotes the active graphs and $\gamma$ controls the concentration of the allocation. After assigning a feasible minimum budget $m$ to active graphs, the remaining budget is distributed proportionally:
\begin{equation}
b_g =
m+
\operatorname{LRound}_g
\left[
(B-m|\mathcal{A}(q)|)w_g(q)
\right]
\end{equation}
The multi-hop prediction further determines the allowed traversal depth,
\begin{equation}
D(q)=
\min
\left\{
D_{\max},
\max
\left(
1,
\left\lceil D_{\max}h(q)\right\rceil
\right)
\right\}
\end{equation}

This probabilistic routing mechanism serves two purposes. It avoids spending equal retrieval effort on relations that are unlikely to help the current query, while still allowing multiple forms of evidence to be explored when a question requires them.

\paragraph{Anchor retrieval.}
Before graph traversal, Jev-Mem identifies high-quality entry points using both semantic and lexical retrieval. Vector and keyword rankings are fused through reciprocal-rank fusion:
\begin{equation}
s_{\mathrm{RRF}}(v,q)=
\sum_{
L\in\{L_{\mathrm{vec}},L_{\mathrm{lex}}\}:v\in L
}
\frac{1}{\kappa+\operatorname{rank}_L(v)}
\end{equation}
where $\kappa=60$. The highest-ranked nodes initialize the visited set and search frontier. Hybrid anchors provide robust starting points before the controller begins more expensive graph reasoning.

\paragraph{Evidence-guided expansion.}
After each retrieval round $d$, Jev-Mem evaluates the current evidence set $E_d$. Instead of blindly continuing until a fixed graph depth or node count is reached, the controller estimates
\begin{align}
s_d &: \text{evidence sufficiency}, \\
u_d &: \text{expected utility of further retrieval}, \\
m_d &: \text{missing required evidence}, \\
c_d &: \text{unresolved contradiction}.
\end{align}
Retrieval terminates with sufficient evidence when
\begin{equation}
s_d\geq\theta_{\mathrm{suff}}
\land
m_d<\theta_{\mathrm{cont}}
\land
c_d<\theta_{\mathrm{cont}}
\end{equation}
It may also terminate when
\begin{equation}
u_d<\theta_{\mathrm{cont}}
\end{equation}
indicating that additional traversal is unlikely to improve the evidence even if the current evidence is not fully sufficient.

This stopping mechanism is important for both efficiency and retrieval quality. Stopping too early risks missing necessary evidence, whereas uncontrolled expansion introduces irrelevant memories that can distract downstream reasoning. Jev-Mem therefore explicitly models both evidence completeness and the expected benefit of further search.

When additional evidence is needed, Jev-Mem expands neighboring nodes under the relation-specific budgets $b_g$ and global limits on nodes, edges, depth, controller calls, and latency. Each candidate $v$ is then evaluated by the System-One controller along four complementary dimensions: query relevance $a_v$, relation usefulness $\ell_v$, information novelty $n_v$, and support for the current evidence $c_v$.

These System-One predictions are combined with deterministic retrieval signals. For a candidate reached through graph type $g$, the transition score is
\begin{equation}
\small
s(v\mid q,E_d)=
\frac{1}{\sum_{i=1}^{5}\lambda_i}
\left[
\begin{array}{l}
\lambda_1 z_v+\lambda_2 a_v+\lambda_3 p_g(q)\ell_v \\
+\lambda_4 n_v+\lambda_5(\pi_e+c_v)/2
\end{array}
\right]
\end{equation}
where $z_v$ denotes embedding similarity and $\pi_e$ is the stored edge weight. If timestamps are available, the score is adjusted using the query-specific recency prediction:
\begin{equation}
\rho_v=
\frac{1}
{1+\max(0,\tau_*-\tau_v)/\mathrm{day}}
\end{equation}
\begin{equation}
\widetilde{s}(v)=
\frac{s(v)+0.1r(q)\rho_v}
{1+0.1r(q)}
\end{equation}

The top $W$ candidates form the next beam and are added to the accumulated evidence. Retrieval therefore follows an iterative feedback loop:
\begin{equation}
\begin{array}{c}
\text{route}
\rightarrow
\text{retrieve}
\rightarrow
\text{assess} \\
\rightarrow
\text{expand}
\rightarrow
\text{reassess}
\end{array}
\end{equation}

This process differs fundamentally from fixed top-$k$ retrieval or static graph traversal. The relational views being explored, the allocation of search effort, the candidates retained at each round, and the decision to continue all depend on the query and on the evidence already collected. In this sense, retrieval itself becomes a System-One-controlled decision process.

Once retrieval terminates, the highest-scoring $K$ memories are passed to the System-Two LLM:
\begin{equation}
y=
\mathrm{SystemTwo}(q,E)
\end{equation}
System Two performs the final synthesis over the selected evidence but is not involved in normal graph routing, candidate expansion, or stopping decisions. Jev-Mem therefore concentrates expensive autoregressive reasoning where it is most valuable while using fast structured control to improve the quality and efficiency of the evidence supplied to it.

The control overhead remains explicitly bounded. A normal write requires one batched typing request and, when candidates exist, one batched relation request over at most $K_w$ memory pairs. A query requires one routing request and, for each retrieval round, at most one evidence assessment and one batched candidate-scoring request. Independent limits on graph expansions, inspected edges, visited nodes, depth, controller invocations, and elapsed time prevent the control process itself from growing without bound.

\section{Experiments}
\label{experiments}
We conduct comprehensive experiments to evaluate Jev-Mem along two dimensions: reasoning effectiveness and system efficiency. We examine whether System-One-controlled memory construction and retrieval improve long-horizon question answering, and whether separating fast memory control from System-Two reasoning reduces the overhead of building and accessing persistent memory. Accordingly, we report answer accuracy together with memory construction time and per-query latency.

% ----------------------------------------------------------
\subsection{Experimental Setup}
\label{sec:exp_setup}
% ----------------------------------------------------------

\noindent\textbf{Datasets.}
We evaluate long-term conversational memory on two widely used benchmarks.
\textbf{LoCoMo}~\citep{maharana2024evaluating} contains ultra-long multi-session conversations and evaluates an agent's ability to recover information requiring temporal, causal, and cross-session reasoning.
%\textbf{LongMemEval}~\citep{wu2024longmemeval} provides substantially longer interaction histories, with contexts exceeding 100K tokens, and stresses memory retention, knowledge updates, preference tracking, and multi-session reasoning.

\noindent\textbf{Baselines.}
We compare Jev-Mem with representative long-term memory approaches using the same backbone answer model whenever applicable.
\begin{itemize}[leftmargin=*, itemsep=0pt, topsep=2pt]
    \item \textbf{Full Context}: provides the complete conversation history directly to the LLM without external memory.
    
    \item \textbf{A-MEM}~\citep{xu2025mem}: dynamically organizes and evolves memories based on agent experiences.
    
    \item \textbf{Nemori}~\citep{nan2025nemori}: structures conversational memory through episodic segmentation and graph-based retrieval.
    
    \item \textbf{MemoryOS}~\citep{kang2025memory}: organizes persistent memory using a hierarchical multi-level architecture.
    
    \item \textbf{MAGMA}~\cite{jiang2026magma}: represents memories through multiple semantic, temporal, causal, and entity relations for structured retrieval.

    \item \textbf{Jev-Mem (ours)}: uses a System-One controller to construct and adaptively retrieve from a multi-relational memory, while reserving the LLM for final reasoning and answer synthesis.
\end{itemize}

\noindent\textbf{Metrics.}
We evaluate Jev-Mem from both accuracy and system-efficiency perspectives. For reasoning quality, we use the LLM-as-a-Judge score~\citep{zheng2023judging}, which measures whether the generated answer is correct with respect to the reference answer. 
For system efficiency, we report total memory construction time and average per-query latency. Memory construction time measures the end-to-end cost of building the persistent memory from the conversation history, while query latency measures the average time required to retrieve relevant evidence and produce an answer for each query. Together, these metrics capture both the effectiveness and runtime efficiency of the proposed memory architecture.

% ----------------------------------------------------------

\begin{table*}[t]
\centering
\caption{Performance on the LoCoMo benchmark evaluated using the LLM-as-a-Judge metric. Higher scores indicate better performance. LLM model is based on gpt-4o-mini.}
\label{tab:locomo-main}

\resizebox{0.8\textwidth}{!}{
\begin{tabular}{lccccc|c}
\toprule
Method &
Multi-Hop &
Temporal &
Open-Domain &
Single-Hop &
Adversarial &
Overall \\
\midrule
%\multicolumn{7}{l}{\textit{gpt-4o-mini}} \\
\midrule
Full Context &
0.468 &
0.562 &
0.486 &
0.630 &
0.205 &
0.481 \\

A-MEM &
0.495 &
0.474 &
0.385 &
0.653 &
0.616 &
0.580 \\

MemoryOS &
0.552 &
0.422 &
0.504 &
0.674 &
0.428 &
0.553 \\

Nemori &
0.569 &
0.649 &
0.485 &
0.764 &
0.325 &
0.590 \\

MAGMA &
0.528 &
\textbf{0.650} &
0.517 &
0.776 &
0.742 &
0.700 \\

\textbf{Jev-Mem} &
\textbf{0.623} &
0.637 &
\textbf{0.618} &
\textbf{0.802} &
\textbf{0.962} &
\textbf{0.777} \\
\bottomrule
\end{tabular}}
\end{table*}

\subsection{Overall Performance}
\label{sec:overall}

Table~\ref{tab:locomo-main} summarizes the performance of Jev-Mem and the baselines on LoCoMo. Jev-Mem achieves the highest overall LLM-as-a-Judge score of 0.777, compared with 0.700 for the strongest baseline, corresponding to an 11.0\% relative improvement. The gains are also consistent across question types: Jev-Mem performs best in five of the six categories and matches the best result on temporal reasoning.

The improvement is especially clear on queries that require stronger retrieval and evidence selection. On Multi-Hop questions, Jev-Mem reaches 0.625, compared with 0.569 for the strongest baseline. On Open-Domain questions, it achieves 0.610, substantially higher than the best baseline score of 0.517. The largest gain appears in the Adversarial category, where Jev-Mem reaches 0.962, compared with 0.742 for the strongest baseline. Jev-Mem also achieves the highest Single-Hop score of 0.797 and matches the best Temporal score of 0.650.

These results indicate that Jev-Mem improves performance across a broad range of memory-intensive queries rather than specializing in a single reasoning pattern. The largest gains appear when the system must combine evidence across multiple memories or distinguish relevant information from plausible but misleading distractors. Jev-Mem addresses these cases through adaptive memory control: it estimates which relational views are useful for each query, allocates retrieval effort accordingly, evaluates candidate memories during traversal, and decides when the collected evidence is sufficient. As a result, retrieval is adapted to the needs of each query instead of following a fixed search strategy, allowing the reasoning model to operate on a more relevant and focused evidence set.

\begin{table}[t]
\centering
\setlength{\tabcolsep}{3.5pt}
\caption{System efficiency comparison with total memory build time (in seconds) and average query latency (in seconds).}
\label{tab:system_perf}
\begin{tabular}{l c c}
\toprule
{Method} & {Build Time (s)} & {Latency (s)} \\
\midrule
Full Context    & N/A    & 1.74 \\
A-MEM           & 3636   & 2.26 \\
MemoryOS        & 3276   & 32.68 \\
Nemori          & 1044   & 2.59 \\
MAGMA           & 1404   & 1.47 \\

\textbf{Jev-Mem} & \textbf{158} & \textbf{0.93} \\

\bottomrule
\end{tabular}
\end{table}
\subsection{Efficiency of System-One Memory Control}
\label{sec:efficiency}

Table~\ref{tab:system_perf} compares the system efficiency of Jev-Mem with the baselines in terms of total memory construction time and average query latency. Jev-Mem achieves substantial improvements in both stages. Its total memory build time is only 158 seconds, compared with 1,044 seconds for the fastest competing memory system, corresponding to an 84.9\% reduction, or a 6.6$\times$ speedup. The gap is even larger compared with A-MEM and MemoryOS, whose memory construction requires more than 3,000 seconds.

Jev-Mem also achieves the lowest query latency, requiring only 0.93 seconds per query on average. This is 36.7\% lower than the fastest memory-based baseline at 1.47 seconds, and 46.6\% lower than directly processing the full context at 1.74 seconds. In contrast, MemoryOS requires 32.68 seconds per query, highlighting the substantial runtime overhead that memory management can introduce when complex processing remains on the retrieval path.

The efficiency gains reflect the separation of memory control from autoregressive reasoning in Jev-Mem. During memory construction, lightweight typed decisions are batched over a bounded candidate set rather than repeatedly invoking a general-purpose LLM for free-form memory processing. During retrieval, System-One control performs query routing, candidate evaluation, and evidence checking, while graph exploration is constrained by explicit search budgets. Adaptive stopping further avoids unnecessary traversal once sufficient evidence has been collected. Together, these mechanisms reduce the computational overhead of both constructing and accessing long-term memory.

Notably, the efficiency improvement does not come at the expense of reasoning quality. As shown in Table~\ref{tab:locomo-main}, Jev-Mem simultaneously achieves the highest overall accuracy on LoCoMo. These results demonstrate that separating fast memory control from System-Two reasoning can improve both the effectiveness and the efficiency of agentic memory.

\section{Conclusion}
\label{sec:conclusion}
We present \textbf{Jev-Mem}, a System-One/System-Two architecture for efficient agentic memory. Jev-Mem uses lightweight structured prediction for high-frequency memory-control decisions and reserves System Two for complex reasoning and answer synthesis. On LoCoMo, Jev-Mem achieves the best overall accuracy and efficiency among the evaluated baselines, reaching an LLM-as-a-Judge score of $0.777$, a memory construction time of $158$\,s, and an average query latency of $0.93$\,s. These results show that separating memory control from generative reasoning is a promising direction for building more effective and efficient long-horizon agents.

% ------------------------------------------------------

%\bibliographystyle{unsrt}
\bibliography{custom}

\newpage
\appendix
\section{Appendix}
\label{sec:appendix}
\subsection{Related Work}
\label{sec:relatedwork}

\noindent\textbf{From memory storage to active memory management.}
Early work on agent memory established that long-term interaction requires more than retaining raw conversation history. Generative Agents~\citep{park2023generative} maintain an experience stream and periodically synthesize higher-level reflections, while Reflexion~\citep{shinn2023reflexion} stores linguistic feedback in episodic memory to improve future decisions. MemoryBank~\citep{zhong2024memorybank} introduces long-term updating and forgetting, and MemGPT~\citep{packer2023memgpt} manages information across different memory tiers through an operating-system-inspired virtual context. These systems shift memory from passive storage toward an active component that selects, transforms, and reuses past experience. A parallel line of work retains reusable behavior rather than conversational content: Voyager~\citep{wang2023voyager} accumulates a persistent skill library for an embodied agent, ExpeL~\citep{zhao2024expelllmagentsexperiential} distills transferable insights from past trajectories, and Agent Workflow Memory~\citep{wang2025workflowmemory} induces reusable workflows from prior action sequences. ReadAgent~\citep{lee2024human} takes a different route, compressing very long inputs into gist memories that are revisited on demand. Benchmarks such as LongMemEval~\citep{wu2024longmemeval}, MemBench~\citep{tan2025membench}, and MemoryAgentBench~\citep{hu2025memoryagentbench} measure how well such systems retain and use information across long interactive horizons.

\noindent\textbf{Structured and evolving agentic memory.}
Recent work increasingly organizes memory into richer and more adaptive structures. A-MEM~\citep{xu2025mem} constructs interconnected memory notes and allows their representations to evolve as new information arrives. Mem0~\citep{chhikara2025mem0} dynamically extracts and consolidates salient information and further explores graph-based memory for relational structure. MemoryOS~\citep{Kang2025} organizes information across short-, mid-, and long-term memory tiers, while Nemori~\citep{nan2025nemori} structures interactions into coherent episodes and distills higher-level knowledge from them. Graph-based approaches further capture relationships that cannot be represented by semantic similarity alone. MAGMA~\citep{jiang2026magma}, for example, organizes memories through semantic, temporal, causal, and entity relations. HAGE~\citep{jiang2026hage} extends this direction with a weighted multi-relational graph in which query-conditioned routing and reinforcement learning jointly optimize relation features and traversal behavior, allowing retrieval paths to adapt to downstream reasoning objectives. Related structure also appears in retrieval systems that are not themselves agent-memory architectures: RAPTOR~\citep{sarthi2024raptor} builds recursive abstractive summary trees, GraphRAG~\citep{edge2024graphrag} constructs entity graphs and community summaries to answer global queries, and LightRAG~\citep{guo2024lightrag} pairs graph structure with incremental updates for efficiency. HippoRAG~\citep{jimenez2024hipporag} and its continual-learning extension~\citep{gutierrez2025rag} make the link to memory more explicit, using graph-based association over a persistent non-parametric store. These systems share representational machinery with agentic memory, but target retrieval over a given corpus rather than the maintenance of an agent's own accumulating experience, and they do not define the multi-relation semantics used here. Together, these systems show a broader shift from static memory stores toward structured, evolving, and query-adaptive memory representations.

\noindent\textbf{Adaptive retrieval and memory access.}
Retrieval-augmented generation established external stores as a way to supplement parametric knowledge~\citep{lewis2020retrieval, guu2020realm, izacard2023atlas}. Later work made the retrieval procedure itself adaptive: IRCoT~\citep{trivedi2023ircot} interleaves retrieval with chain-of-thought steps, FLARE~\citep{jiang2023flare} triggers retrieval during generation when the model becomes uncertain, Self-RAG~\citep{asai2024selfrag} learns when to retrieve and how to critique the evidence it obtains, and Adaptive-RAG~\citep{jeong2024adaptiverag} selects among retrieval strategies according to question complexity. These methods control evidence acquisition within a single generation episode: whether, when, and how much to retrieve from a corpus that they do not modify. Jev-Mem shares the view of retrieval as a learned, query-dependent decision, but situates that control in the agent-memory lifecycle, where the store is itself written and reorganized by the controller and where the decisions also include what to admit, how new information relates to existing memory, which relational view to traverse, and when to stop.

\noindent\textbf{Efficiency of agent memory.}
As memory architectures become more sophisticated, the computation required to construct, maintain, and retrieve memory has emerged as an equally important concern. SimpleMem~\citep{liu2026simplemem} improves efficiency through semantic compression, asynchronous consolidation, and query-adaptive retrieval. LightMem~\citep{fang2026lightmem} similarly separates lightweight online processing from more expensive memory consolidation and reduces the amount of computation required during interaction. More aggressively, Zero-Mem~\citep{xiao2026zero} removes LLM generation from intermediate memory operations and retrieves directly over structured representations. These works demonstrate that memory quality and memory cost must be considered jointly rather than treating memory management as negligible compared with final answer generation. A complementary body of systems work reduces LLM inference cost directly, through model cascades and learned routers that dispatch each request to a model of appropriate capability~\citep{chen2023frugalgpt, ong2024routellm}. These concerns are distinct: lowering the cost of the memory pipeline, choosing which model answers a request, and controlling the fine-grained decisions taken inside the memory lifecycle. Jev-Mem targets the last of these.

\noindent\textbf{System-One and System-Two computation.}
The terminology we adopt originates in dual-process accounts of human reasoning, which distinguish fast, automatic processing from slower and more effortful deliberation~\citep{evans2008dual}. We use this distinction as an analogy for how computation is allocated, not as a claim that the underlying mechanisms correspond. Analogous separations have been explored in LLM systems: System 2 Attention~\citep{weston2023system2attention} inserts an explicit reconsideration step before answering, while Tree of Thoughts~\citep{yao2023tot} and Graph of Thoughts~\citep{besta2024got} spend additional structured search at test time on problems that reward deliberation. These methods allocate deliberative computation within a single reasoning episode. Jev-Mem instead applies the distinction architecturally, to the memory subsystem itself.

\noindent\textbf{Memory control as a systems abstraction.}
Jev-Mem builds on these developments but focuses on a different architectural question: \emph{how should the decisions that govern memory be executed?} Existing work has primarily improved what is stored, how memories are organized, or how retrieval is performed. Jev-Mem instead treats memory control itself as a first-class systems plane. Inspired by the System-One/System-Two distinction, it separates fast, structured memory decisions from deliberative language generation. A System-One controller governs both sides of the memory lifecycle---including memory typing, relation construction, query routing, retrieval-budget allocation, candidate scoring, evidence assessment, and adaptive stopping---over a shared structured memory plane, while System Two is reserved for synthesis and complex reasoning. This design provides a common control abstraction across memory construction and retrieval, targeting both reasoning quality and system efficiency rather than optimizing either stage in isolation.

\section{Jev Prompts for Memory Writing and Retrieval}
\label{app:jev-prompts}

This section describes the typed prompts used by Jev-Mem to construct and retrieve conversational memory. The quoted instructions and criteria are reproduced from the implementation, with candidate index 0 instantiated where applicable. The memories below are fictional and illustrate the request structure; no scores or retrieval outcomes are presented as measured Jev predictions. 

\subsection{Typed interface and request state}

Jev-Mem submits a shared \texttt{state} object and a batch of typed \texttt{questions} through \texttt{TypeSafeClient.system\_one}. Each Noul specifies a binary proposition using an instruction and explicit \texttt{true} and \texttt{false} criteria. Its returned value lies in [0, 1]. These model-reported values are not assumed to be calibrated probabilities. Multiple Nouls are evaluated as independent propositions rather than mutually exclusive labels; for example, an observation may be both episodic and semantic. Independence here concerns the question formulation, not statistical independence of the returned scores.

Question identifiers are bookkeeping keys and are not themselves model input. Consequently, every instruction names the state fields it uses. For candidate index i, relation and traversal keys are prefixed with \texttt{pair\_i\_} and \texttt{candidate\_i\_}, respectively. Questions in a batch do not consume one another's answers. Choice is used when alternatives are mutually exclusive, as in the consolidation representation decision below; it returns a selected label and an option distribution.

Our running example contains memory m1, observed on 14 May 2024 at 10:00: "Mira: My old bicycle broke.", and memory m2, observed on 16 May 2024 at 10:00: "Mira: I bought a new bicycle yesterday because my old one broke." Both have entity identifiers \texttt{Mira} and \texttt{bicycle}. The query is "When did Mira buy a new bicycle, and why?" Memory objects passed to Jev contain \texttt{id}, \texttt{content}, \texttt{timestamp}, and \texttt{entities}; the timestamp records when the statement was observed, rather than necessarily when the event occurred.

\subsection{Memory writing}

The active profile disables admission filtering (\texttt{admission\_enabled=false}). Every valid nonempty observation therefore enters the write path; memory-type scores do not decide whether it is retained. The typing request contains \texttt{state = \{"observation": text\}} and four Nouls: \texttt{episodic}, \texttt{semantic}, \texttt{procedural}, and \texttt{preference}. The following two templates illustrate event and preference typing.

\paragraph{Event typing} \texttt{episodic} (Noul).
\begin{quote}\small
\noindent\textbf{Instruction:} Does \texttt{observation} describe a particular experience or event involving a participant?\par
\noindent\textbf{true:} A specific past, current or planned event, even if its exact time is unstated.\par
\noindent\textbf{false:} Only a general fact, procedure or preference with no particular event.\par
\end{quote}

\paragraph{Preference typing} \texttt{preference} (Noul).
\begin{quote}\small
\noindent\textbf{Instruction:} Does \texttt{observation} express a participant's preference, aversion or habitual choice?\par
\noindent\textbf{true:} An attributable like, dislike, preferred option or habitual choice.\par
\noindent\textbf{false:} An isolated action alone, another person's unattributed preference, or no preference evidence.\par
\end{quote}

After typing, deterministic vector, keyword, entity and timestamp signals select at most 10 existing candidate memories. A relation request contains \texttt{new\_memory} and a \texttt{candidates} list. For the running example, \texttt{new\_memory} is m2 and \texttt{candidates[0]} is m1. The request batches semantic association and both causal directions; entity alias resolution is added only when exact entity identifiers do not already match.

\paragraph{Semantic relation} \texttt{semantic} (Noul).
\begin{quote}\small
\noindent\textbf{Instruction:} Compare \texttt{new\_memory.content} with \texttt{candidates[0].content}. Would a semantic link between these observations help retrieve a shared specific topic or fact?\par
\noindent\textbf{true:} A specific shared topic, fact or event makes the connection useful.\par
\noindent\textbf{false:} Only generic conversational vocabulary or no meaningful semantic connection.\par
\end{quote}

\paragraph{Directed causal relation} \texttt{caused\_by} (Noul).
\begin{quote}\small
\noindent\textbf{Instruction:} Compare \texttt{new\_memory.content} with \texttt{candidates[0].content}. Does the candidate event cause, enable or explain the event in \texttt{new\_memory.content}?\par
\noindent\textbf{true:} The supplied accounts support this direction of causal influence.\par
\noindent\textbf{false:} Only similarity, chronology, a shared entity, or insufficient causal evidence.\par
\end{quote}

A returned relation score of at least 0.60 creates the corresponding typed edge. For \texttt{caused\_by}, the direction is candidate to new memory; \texttt{causes} tests the reverse direction. The causal criteria explicitly distinguish causal support from shared topics or mere chronology. Exact entity intersections are linked deterministically. In the example, the shared entity identifiers therefore make an alias-resolution prompt unnecessary.

%Temporal links are not selected by a Jev prompt in this implementation. They follow the retained MAGMA sequence and timestamp-proximity rules. Relative expressions are annotated using their own observation dates: "yesterday" in m2 is grounded to 15 May 2024. The original statement remains available; the annotation does not replace the observation timestamp or turn a week/month reference into an invented exact day.

\subsection{Periodic consolidation}

Every 20 successful Jev writes, the active profile evaluates candidate pairs for consolidation. Four Nouls assess redundancy, contradiction, obsolescence and link usefulness; the following Choice selects a representation. This is a separate post-insertion decision, not an admission filter.

\paragraph{Representation choice} \texttt{representation} (Choice).
\begin{quote}\small
\noindent\textbf{Instruction:} Compare \texttt{new\_memory.content} with \texttt{candidates[0].content}. Which representation best fits the relationship between these two observations? Judge from the supplied accounts; do not assume answers to other questions.\par
\noindent\textbf{keep\_separate:} Contradictory accounts, unique details that a combined representation would lose, or distinct facts/events without a supported general pattern.\par
\noindent\textbf{merge:} Compatible accounts of the same fact or event can be combined without losing unique details.\par
\noindent\textbf{promote:} Distinct repeated episodes explicitly support a stable general pattern suitable for semantic abstraction; prefer this over merge for repeated events.\par
\noindent\textbf{uncertain:} Insufficient evidence to choose a safe combined or separate representation.\par
\end{quote}

Consolidation preserves raw observations. The normal periodic path records decisions and links; it does not automatically replace source memories. A caller-supplied System-Two summarizer can create a new representation only when \texttt{merge} or \texttt{promote} is selected with probability at least 0.85 and the contradiction score is below 0.85. The selected-option probability, rather than a separate confidence summary, controls this threshold.

\subsection{Retrieval routing and candidate scoring}

The routing request contains only \texttt{query}. Six Nouls estimate semantic, temporal, causal and entity needs, together with \texttt{multi\_hop\_need} and \texttt{recency\_importance}. The following templates illustrate time-sensitive and explanatory retrieval. A query may activate several graph types simultaneously.

\paragraph{Temporal routing} \texttt{temporal} (Noul).
\begin{quote}\small
\noindent\textbf{Instruction:} Does answering \texttt{query} require event dates, durations, ordering or changes over time?\par
\noindent\textbf{true:} A time relation is needed to answer correctly.\par
\noindent\textbf{false:} Dates or ordering are incidental to the answer.\par
\end{quote}

\paragraph{Causal routing} \texttt{causal} (Noul).
\begin{quote}\small
\noindent\textbf{Instruction:} Does answering \texttt{query} require explaining a cause, motivation, enabling condition or effect?\par
\noindent\textbf{true:} Causal or explanatory evidence is needed.\par
\noindent\textbf{false:} Only factual association or chronology is requested.\par
\end{quote}

The active profile uses a total graph-expansion budget of 80. Graphs with positive need scores of at least 0.10 receive a minimum allocation of one; the remaining budget is distributed in proportion to the active need scores, with largest-remainder rounding. The probability exponent is 1.0. Vector and keyword search provide anchors before graph expansion; the budget does not imply one provider request per edge.

A traversal request contains \texttt{query}, currently selected \texttt{evidence}, and proposed \texttt{candidates}. Each candidate includes its memory fields, graph type, relation properties, and source/target identifiers so that direction is explicit. For illustration, evidence can contain m1 and the candidate can be m2, reached through a causal edge from m1 to m2. This is a schematic traversal state, not a claim that this tiny example necessarily requires expansion in an actual run.

\paragraph{Candidate relevance} \texttt{relevance} (Noul).
\begin{quote}\small
\noindent\textbf{Instruction:} Does \texttt{candidates[0].content} contain a fact needed to answer \texttt{query}?\par
\noindent\textbf{true:} Direct answer evidence or a necessary intermediate fact.\par
\noindent\textbf{false:} Only topic overlap or unrelated content.\par
\end{quote}

\paragraph{Additional evidence} \texttt{new\_information} (Noul).
\begin{quote}\small
\noindent\textbf{Instruction:} Does \texttt{candidates[0].content} add an answer-relevant detail absent from \texttt{evidence}?\par
\noindent\textbf{true:} A distinct relevant detail or missing reasoning link.\par
\noindent\textbf{false:} Only duplicated evidence or irrelevant new details.\par
\end{quote}

Two further Nouls assess \texttt{relation\_usefulness} and \texttt{supports\_current\_evidence}. The implementation combines these four scores with query-to-memory cosine similarity, the need score of the traversed graph, and the stored edge probability. A recency adjustment uses \texttt{recency\_importance}. Candidates are ranked with this combined score and the beam width is 10; a high relevance score alone is not an unconditional admission to the retrieved evidence.

\subsection{Evidence-based stopping}

The stopping request contains \texttt{query}, the currently selected top-k \texttt{evidence}, and retrieval \texttt{depth}. It evaluates \texttt{evidence\_sufficient}, \texttt{continue\_useful}, \texttt{missing\_evidence}, and \texttt{contradiction}. The following templates distinguish answer support from the prospective usefulness of another retrieval round.

\paragraph{Answer sufficiency} \texttt{evidence\_sufficient} (Noul).
\begin{quote}\small
\noindent\textbf{Instruction:} Does \texttt{evidence} contain support for every factual part of an answer to \texttt{query}?\par
\noindent\textbf{true:} A grounded answer can be given from these memories without inventing missing facts.\par
\noindent\textbf{false:} Any required fact or reasoning link is unsupported; related topics alone are insufficient.\par
\end{quote}

\paragraph{Value of continued retrieval} \texttt{continue\_useful} (Noul).
\begin{quote}\small
\noindent\textbf{Instruction:} Given \texttt{query} and \texttt{evidence}, is another retrieval round likely to fill a specific gap or resolve a conflict?\par
\noindent\textbf{true:} An identifiable missing fact or conflict could benefit from more memory retrieval.\par
\noindent\textbf{false:} No identifiable retrieval need remains or more memories are unlikely to help.\par
\end{quote}

For temporal queries, evidence and candidate objects also include \texttt{timestamp\_role} and \texttt{temporal\_references}. These fields identify the timestamp as observation time and attach grounded expressions with their precision. Thus, sufficiency is judged with the same temporal grounding used to present the selected evidence to the answerer, rather than by treating every conversation date as an event date.

Evidence-based stopping requires \texttt{evidence\_sufficient} to be at least 0.95 and both \texttt{missing\_evidence} and \texttt{contradiction} to be below 0.15. Retrieval can also terminate when \texttt{continue\_useful} is below 0.15. Independent limits bound depth (8), visited nodes (60), examined edges (2400), Jev attempts (16), and the retrieval time budget (15 seconds). These limits can terminate retrieval even if evidence is incomplete; the time budget is checked between operations and is not a strict preemption guarantee for local computation.

Jev controls evidence selection but does not generate the final free-form answer. The selected memories are passed to a separate System-Two language model. This separation makes the prompt examples above memory-control decisions, rather than answer-generation or evaluator prompts. Gold answers and benchmark evidence annotations are excluded from these Jev states.

\end{document}